\pdfoutput=1

\documentclass[11pt]{article}

\usepackage[]{EMNLP2022}

\usepackage{times}
\usepackage{latexsym}
\usepackage{graphicx}
\usepackage{stfloats}
\usepackage{enumitem}
\usepackage{booktabs}
\usepackage{multirow}
\usepackage{subfigure}
\usepackage{hyperref}
\usepackage[T1]{fontenc}

\usepackage[utf8]{inputenc}

\usepackage{microtype}

\usepackage{inconsolata}

\title{Improving Chinese Named Entity Recognition by Search Engine Augmentation}

\author{Qinghua Mao \and Kui Meng \\
  Shanghai Jiao Tong University \\
  \texttt{\{mmmm2018,mengkui\}@sjtu.edu.cn} \\\And
  Jiatong Li \\
  The University of Melbourne \\
  \texttt{jiatongl3@student.unimelb.edu.au}
  }

\begin{document}
\maketitle
\begin{abstract}
Compared with English, Chinese suffers from more grammatical ambiguities, like fuzzy word boundaries and polysemous words. In this case, contextual information is not sufficient to support Chinese named entity recognition (NER), especially for rare and emerging named entities. Semantic augmentation using external knowledge is a potential way to alleviate this problem, while how to obtain and leverage external knowledge for the NER task remains a challenge. In this paper, we propose a neural-based approach to perform semantic augmentation using external knowledge from search engine for Chinese NER. In particular, a multi-channel semantic fusion model is adopted to generate the augmented input representations, which aggregates external related texts retrieved from the search engine. Experiments have shown the superiority of our model across 4 NER datasets, including formal and social media language contexts, which further prove the effectiveness of our approach.
\end{abstract}

\section{Introduction}
Different from English, Chinese is correlated with word segmentation and suffers from more polysemous words and grammatical ambiguities. Given that contextual information is limited, external knowledge is leveraged to support the entity disambiguation, which is critical to improve Chinese NER, especially for rare and emerging named entities. \par
\begin{figure}[ht]
\centering
\includegraphics[scale=0.225]{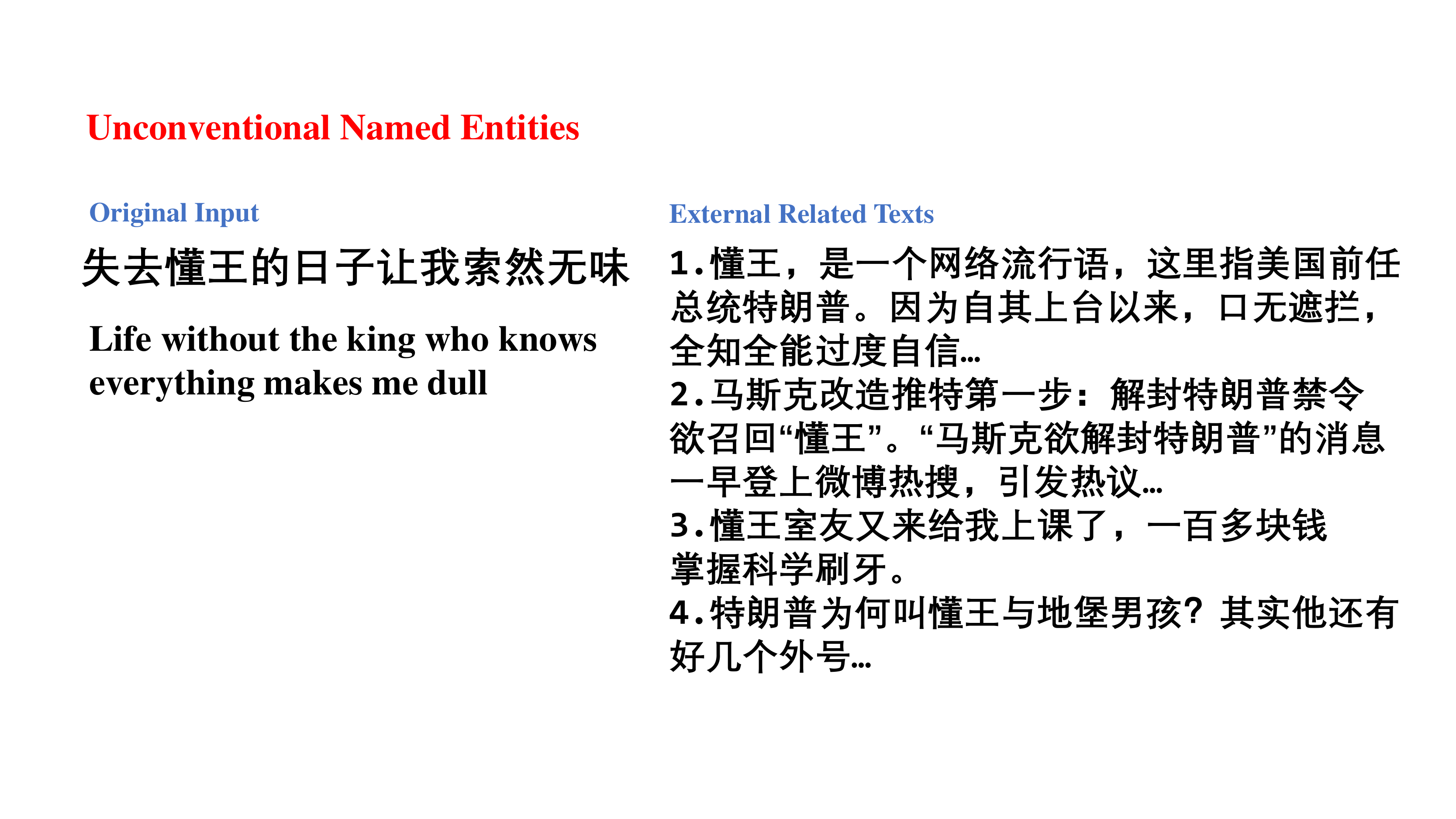}
\caption{A motivating example for recognizing new entities using external related texts retrived from the search engine, in which "the king knows everything"("Dong Wang")is an unconventional named entity referring to Donald J. Trump.}
\label{fig:new_example}
\end{figure}

Apart from lexical information \cite{gui-etal-2019-lexicon}, other external sources of information has been leveraged to perform semantic augmentation for NER, such as external syntactic features \cite{li2020wcp}, character radical features \cite{xu2019exploiting} and domain-specific knowledge \cite{zafarian2019improving}. However, it takes extra efforts to extract these information and most of them are domain-specific. Search engine is a straightforward way to retrieve open-domain external knowledge, which can be evidence for recognizing those ambiguous named entities. A motivating example is shown in Figure \ref{fig:new_example}. \par
In this paper, we suggest to improve Chinese NER by semantic augmentation through a search engine. Inspired by Fusion-in-Decoder \cite{izacard2021leveraging}, we propose a multi-channel semantic fusion NER model which leverages external knowledge to augment the contextual information of the original input. Given external related texts retrieved from the search engine, our model first adopts multi-channel BERT encoder to encode each texts independently. An attention fusion layer is utilized to incorporate external knowledge into the original input representation. Finally, the fused semantic representation is fed into CRF layer for decoding. \par
We implement an external related texts generation module for optimizing retrieval results from the search engine. TextRank \cite{mihalcea2004textrank} and BM25 \cite{robertson2009probabilistic} are utilized to generate external related texts which are semantically relevant to the original input sentence.\par
The experimental results on the generic domain and social media domain show the superiority of our approach. It demonstrates that search engine augmentation can effectively improve Chinese NER, especially for the social media domain.

\section{Model}
The proposed approach can be described in two steps. Given an input sentence, external related texts are retrieved from a search engine. The original input sentence, along with external related texts, is fed into the multi-channel semantic fusion NER model to generate fused representations which aggregates external knowledge obtained from the search engine.
\begin{figure*}[ht]
\centering
\includegraphics[width=\textwidth]{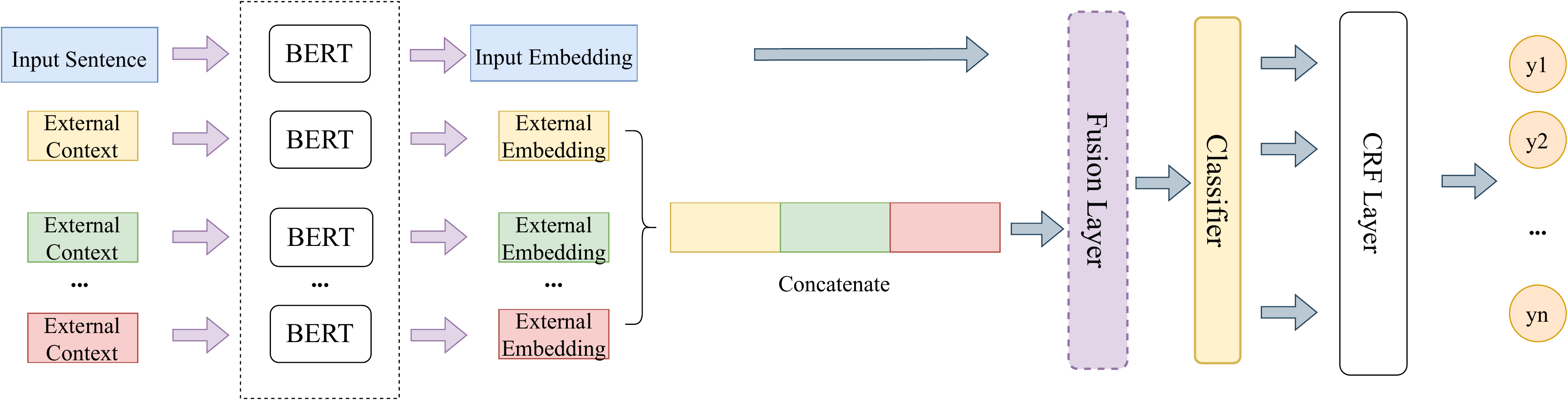}
\caption{The multi-channel semantic fusion NER model. Multi-channel BERT encodes each text independently. Fusion layer generates fused semantic representations based on the attention mechanism, which are fed into CRF layer for named entity prediction.}
\label{fig:model}
\end{figure*}

\subsection{Multi-channel Semantic Fusion\label{sec:fusion}}
We view NER task as a sequence labeling problem. Our multi-channel semantic fusion NER model is shown in Figure \ref{fig:model}, in which BERT-CRF \cite{souza2019portuguese} serves as the backbone structure. \par
Given original input $x$ and $K$ external related texts $\tilde{X}=\{\tilde{x_1},\tilde{x_2},...,\tilde{x_K}\}$, multi-channel BERT encoder is utilized to encode each texts independently, from which original input embedding $H_x$ and external embedding $H_{external}$ are obtained.   
\begin{equation}
    [H_{x},H_{external}]=BERT([x,\tilde{X}])\label{con:encode}
\end{equation} 
where $H_{external}=\{h_{\tilde{x_1}},h_{\tilde{x_2}},...,h_{\tilde{x_K}}\}$. \par
Processing texts independently with a multi-channel encoder means that the computation time grows linearly as the number of texts scales. So it makes the model more extensible. Meanwhile, contextual information of each channel is independent to facilitate the subsequent semantic fusion.\par
\begin{figure}[ht]
\centering
\includegraphics[scale=0.22]{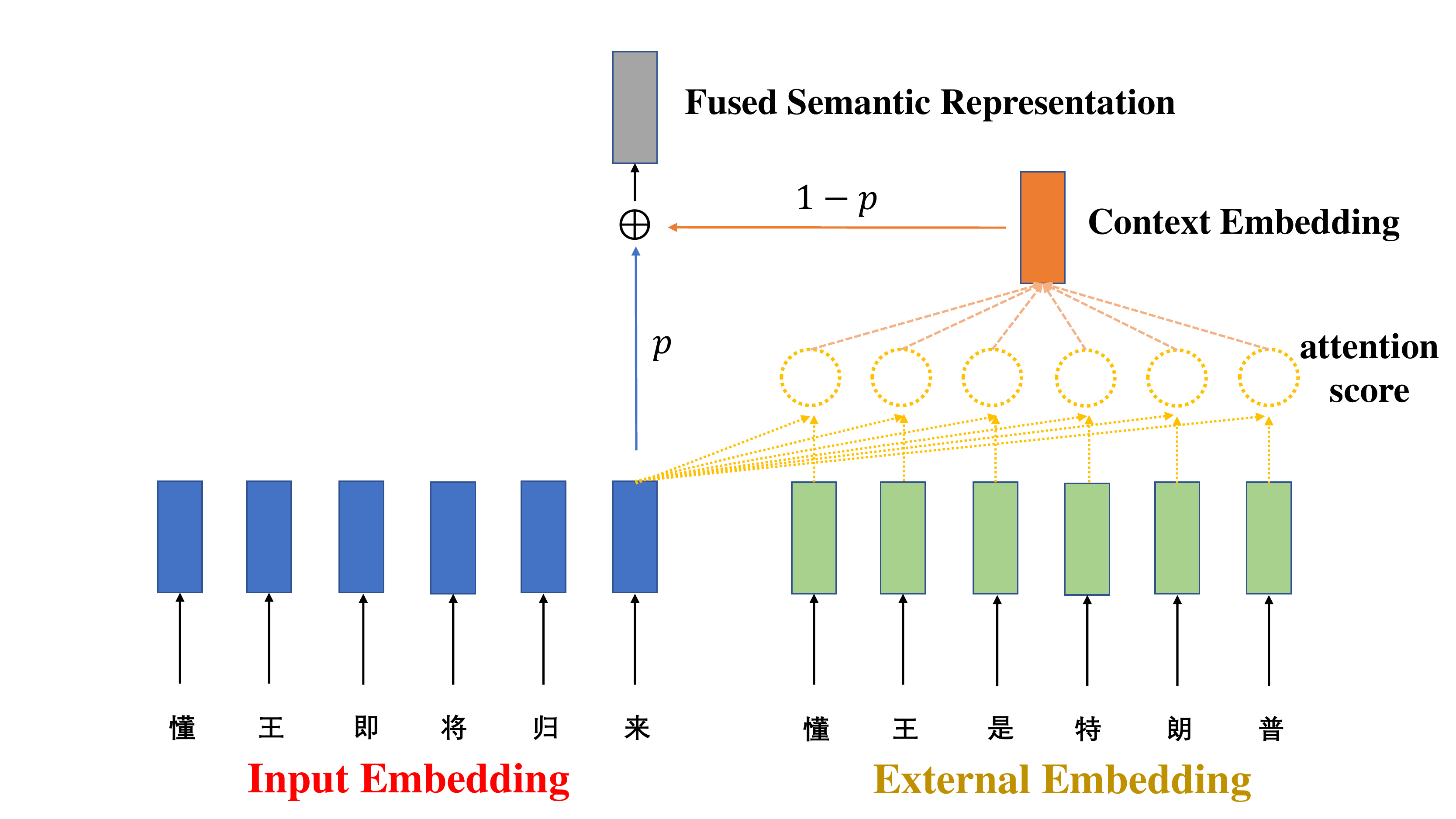}
\caption{A token-level illustration of semantic fusion.}
\label{fig:attnfusion}
\end{figure}
We feed input embedding and context embedding into the attention fusion layer to generate fused semantic representation, which is finally put into CRF layer for decoding. In particular, for input embedding $H_x$, we compute attention scores over external embedding $H_{external}$ to generate context embedding $H_{context}$, which fuses external knowledge based on semantic relevance to the original input. The fused semantic representation $H_{fusion}$ is acquired by calculating the weighted sum of input embedding and context embedding. The weights of input embedding and context embedding are set to a fusion factor $p$ and $1-p$ respectively. A token-level illustration is shown in Figure \ref{fig:attnfusion}.
\begin{equation}
    H_{context} = Attention(H_x,H_{external})\label{con:attention}
\end{equation}
\begin{equation}
    H_{fusion}=p \times H_x+(1-p) \times H_{context}\label{con:fusion_factor}
\end{equation}\par
Three points are taken into account when designing the fusion layer. First, sequence dependency should be reserved, which is very important for NER. Secondly, relation between original input and external contexts should be considered in the semantic fusion, i.e., the former is given priority and external knowledge is supplementary evidence. Thirdly, not all external related texts are necessary for semantic augmentation, we should focus on that part which is of help to accurately identify the named entity.\par

\begin{table*}[!hpt]
  \begin{center}
  \resizebox{\textwidth}{!}{%
  \begin{tabular}{ccccccccccccc}
  \hline
  \multirow{3}{*}{Models} & \multicolumn{6}{c}{Generic NER} & \multicolumn{6}{c}{Social Media NER} \\  \cline{2-13}
  & \multicolumn{3}{c}{Chinese Resume} & \multicolumn{3}{c}{People's Daily} & \multicolumn{3}{c}{Weibo NER} & \multicolumn{3}{c}{Unconventional} \\  \cline{2-13}
          & Precision & Recall & F1     & Precision & Recall & F1     & Precision & Recall & F1     & Precision & Recall & F1     \\ \hline
BERT-CRF  & 0.9549    & 0.9607 & 0.9578 & 0.9178    & \textbf{0.9142} & 0.9160 & \textbf{0.7189}    & 0.6584 & 0.6873 & 0.8320    & 0.8094 & 0.8205 \\ \hline
FLAT+BERT & 0.9540    & 0.9552 & 0.9546 & 0.8688    & 0.9128 & 0.8902 & 0.6794    & 0.6388 & 0.6584 & 0.7807    & 0.7969 & 0.7887 \\ \hline
Ours      & \textbf{0.9618}    & \textbf{0.9648} & \textbf{0.9633} & \textbf{0.9243}    & 0.9126 & \textbf{0.9184} & 0.7121    & \textbf{0.7050} & \textbf{0.7085} & \textbf{0.8530}    & \textbf{0.8298} & \textbf{0.8412} \\ \hline
\end{tabular}%
}
  \end{center}
  \caption{A comparison of the proposed model and other two models. The result shows the superiority of our model.}
  \label{tab:comparison} 
\end{table*}

\subsection{External Knowledge Retrieval}
Our approach retrieves external knowledge by an off-the-shelf search engine. The search engine has been optimized for a fast speed as first priority, which may produce some inaccurate results. Therefore, we implement an generation module to obtain external related texts semantically similar to the original input.\par 
\begin{figure}[ht]
\centering
\includegraphics[scale=0.3]{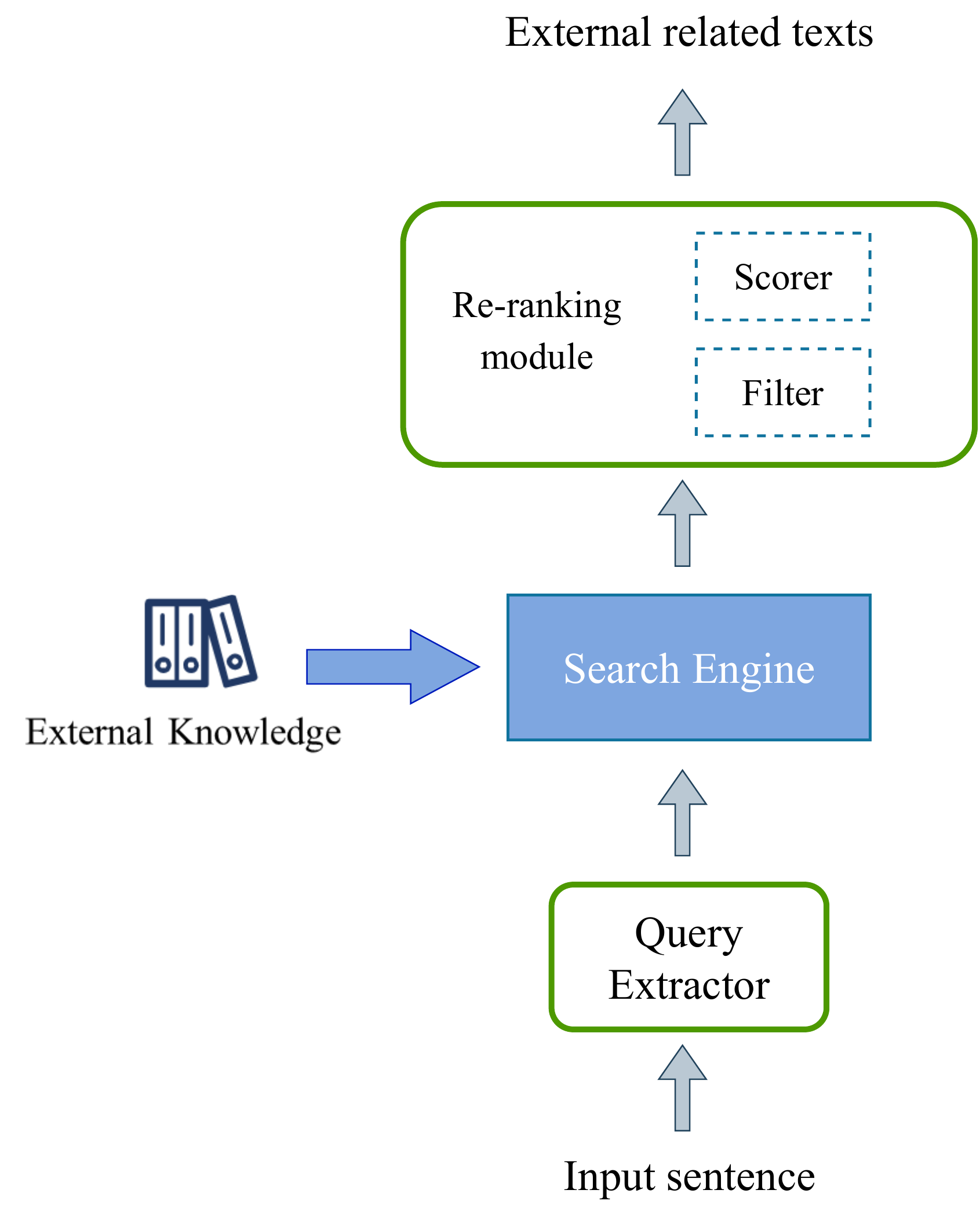}
\caption{The pipeline of external knowledge retrieval through a search engine, which consists of query keywords generation, search engine retrieval and semantic similarity re-ranking three parts.}
\label{fig:search}
\end{figure}
The pipeline of retrieving external related texts in show in Figure \ref{fig:search}. TextRank algorithm is used to generate $m$ query keywords from original input sentence, which are fed into the search engine. We adopt BM25 algorithm to re-rank the retrieved texts according to their semantic similarity to the original input sentence and choose top-K texts as the external related texts. TextRank is a graph-based algorithm which uses certain ways to calculate relation between words and select high ranked words as query keywords. BM25 is a widely-used text similarity algorithm based on term frequency appeared in the corpus. These two statistic-based unsupervised approaches utilize word frequency to calculate semantic relevance, rather than semantic generation using word embedding, in order to prevent information leakage of named entities.    
\section{Experiments}
\subsection{Experimental Setup}
We evalutate our model on four Chinese NER datasets, including
Resume NER \cite{zhang-yang-2018-chinese}, People's Daily\footnote{\url{https://icl.pku.edu.cn/}}, Weibo NER \cite{peng-dredze-2015-named} and our self-built dataset mostly containing unconventional named entities from Chinese social media, e.g, weibo and bilibili\footnote{\url{https://www.bilibili.com/}}. A detailed description can be found in \nameref{sec:appendix}.

\subsection{Overall Performance}

We compare our model with BERT-CRF \cite{souza2019portuguese} and Flat Lattice Transformer \cite{zhang-yang-2018-chinese}, in which BERT-CRF is baseline model. Table \ref{tab:comparison} shows the overall performance. It is remarkable that our model achieves the best performance across 4 Chinese NER datasets with the highest F1 scores. For Chinese NER in generic domain, F1 score of our model achieves 96.33\% on \textbf{Chinese Resume} and 91.84\% on \textbf{People's Daily}, improving by about 0.55\% and 0.24\% respectively comparing to BERT-CRF baseline. For Chinese NER in the social media domain, F1 score of our model achieves 70.85\% on \textbf{Weibo NER} and 84.12\% on our \textbf{Unconventional Named Entity} dataset, improving by about 2.12\% and 2.07\% respectively comparing to BERT-CRF baseline. \par
Compared with formal text, more emerging and rare named entities exist in a social media text, with short text lengths and prominent grammatical irregularities. Limited contextual information within the input makes it harder for social media NER. As can be seen in Table \ref{tab:comparison}, a big gap still remains between generic NER and social media NER. \par 
Furthermore, It is obvious that the improvement of F1 score for social media NER is greater than that for generic NER, which indicates that search engine augmentation is of greater help to enrich the contextual information of social media texts. For rare and unconventional named entities, external knowledge retrieved from the search engine can provide convincing evidence to support named entity disambiguation.
\subsection{Effect of Fusion Strategies}

We compare two fusion strategies, including a linear fusion layer and an attention fusion layer. F1 scores over four datasets are shown in Table \ref{tab:compare_fusion}. F1 score for attention fusion layer is higher than that for linear fusion layer across all datasets. Besides, F1 score for linear fusion is even lower than that for BERT-CRF baseline in Table\ref{tab:comparison}. In particular, the decline of F1 score in the social media domain (8.24\% on \textbf{Weibo NER} and 3.07\% on \textbf{Unconventional Named Entity} dataset) is greater than that in generic domain (1.01\% on \textbf{Chinese Resume} and 2.90\% on \textbf{People's Daily}).\par
The result indicates that linear fusion is not a good choice to perform semantic augmentation for Chinese NER. A linear transformation treats original input and external knowledge equally and does not take token sequence order into consideration. In some way, the generated vector makes no sense. By contrast, attention fusion can better handle sequence dependency and aggregate external knowledge which are more conductive for NER task.  

\begin{table*}[!hpt]
\begin{center}
\begin{tabular}{ccccc}
\hline
      & Chinese Resume & People's Daily & Weibo NER & Unconventional \\ \hline
Linear    & 0.9477         & 0.8870         & 0.6049    & 0.7898         \\ \hline
Attention & 0.9633         & 0.9184         & 0.7085    & 0.8412        \\ \hline
\end{tabular}
\end{center}
\caption{F1 score of two fusion strategies. Linear refers to applying a linear fusion layer while Attention refers to applying an attention fusion layer. An illustration of fusion layers can be found in the supplementary material.}
  \label{tab:compare_fusion} 
\end{table*}

\section{Related Work}
Named entity recognition(NER) has been solved as a sequence labeling problem. Different from English, Chinese is correlated with word segmentation, which means Chinese NER may suffer from fuzzy word boundaries and grammatical irregularities, so character-based models have been the dominant approaches. To fully leverage word-level information, word-character lattice structure \cite{zhang-yang-2018-chinese,li-etal-2020-flat} are utilized to incorporate latent word information into character-based models.\par
Supervised NER approaches usually perform well on specific domains but sometimes fail to recognize rare and emerging named entities. For compensation, external knowledge has been leveraged to improve NER performance. In particular, lexicon features \cite{gui-etal-2019-lexicon,sui-etal-2019-leverage,ma-etal-2020-simplify} have been widely used to enrich word-level information. Some previous works exploit other external sources of information as a feature to formulate hybrid representations, e.g., POS tags \cite{cai2019deep,li2020wcp}, character radical features \cite{xu2019exploiting}, named entity gazetteers \cite{yamada-etal-2020-luke,zafarian2019improving}, etc. \par 
Search engine is a straightforward and effective way to obtain external knowledge. \citet{wang-etal-2021-improving} propose to improve NER by retrieving related contexts from a search engine. They implement a transformer-based model to fuse input sentence and external texts together. By contrast, we utilize a multi-channel BERT to encode each text independently and an attention fusion layer to aggregate external knowledge, which is more suitable for handling large number of retrieved contexts.

\section{Conclusion}
In this paper, we propose a neural-based approach to perform semantic augmentation for Chinese NER. We utilize a search engine to retrieve external related texts which are semantically relevant to the original input, then feed them into the multi-channel semantic fusion NER model to generate fused semantic representation, which aggregates external knowledge as evidence. Empirical results over 4 datasets show our approach is competitive with the baseline and previous state-of-the-art models. The result proves the effectiveness of search engine augmentation, especially for social media NER.

\section*{Limitations}
We propose an innovative idea to improve the semantic representations for Chinese NER using external knowledge from a search engine. Due to time constraints, our experiments still need to be improved to support the effectiveness of external knowledge augmentation in this paper. First, so far we only implement an attention fusion layer and a linear fusion layer for comparison, more semantic fusion approaches need to be improved. Second, query keywords extraction and text similarity re-ranking module can be carefully implemented using other approaches. Third, our external knowledge augmentation approach is applied for token-level sequence labeling using input representations. With the development of seq2seq generative models, external knowledge may be helpful to guide the generation of named entities appeared in the sentence, which is to be discussed.

\section*{Acknowledgements}
We would like to thank the anonymous reviewers for their insightful comments.

\bibliography{anthology,custom}
\bibliographystyle{acl_natbib}

\section*{Appendix \label{sec:appendix}}

\subsection*{Datasets}
To evaluate the effectiveness of our approach, we conducted a series of experiments on 4 Chinese NER datasets:
\begin{itemize}[labelsep = .5em, leftmargin = 0pt, itemindent = 1em]
    \item \textbf{Generic NER:} We use Chinese Resume and People's Daily, two public datasets of generic named entities on formal text.
    \item \textbf{Social Media NER:} We use weibo NER as public dataset of generic named entities on social media. Besides, we build a dataset of unconventional named entities.
\end{itemize}
\par
Unconventional named entities in social media text are transformed by generic named entities through abbreviations, homonyms and feature reference,etc. We categorize them into six classes: Person, Location, Organization, Event, Object, Slogan.\par
Statistics of four datasets are shown in Table \ref{tab:dataset}.
\begin{table}[!hpt]
  \begin{center}
  
    \begin{tabular}{cccc} 
    \hline
    \multicolumn{4}{c}{\textbf{Chinese Resume}} \\ \hline
      & Sentences & Characters &  Entities\\
      \hline
      Train Set & 3821 & 124099 & 13343 \\
      Dev Set & 463 & 13890 & 1488 \\
      Test Set & 477 & 15100 & 1630 \\ \hline
      \multicolumn{4}{c}{\textbf{People's Daily}} \\ \hline
      & Sentences & Characters &  Entities\\
      \hline
      Train Set & 20864 & 979180 & 33992 \\
      Dev Set & 2318 & 109870 & 3819 \\
      Test Set & 4636 & 219197 & 7707 \\  \hline
      \multicolumn{4}{c}{\textbf{Weibo NER}} \\ \hline
      & Sentences & Characters &  Entities\\
      \hline
      Train Set & 1350 & 73778 & 1885 \\
      Dev Set & 270 & 14509 & 389 \\
      Test Set & 270 & 14842 & 414 \\ \hline
      \multicolumn{4}{c}{\textbf{Unconventional}} \\ \hline
      & Sentences & Characters &  Entities\\
      \hline
      Train Set & 2344 & 143257 & 6211 \\
      Dev Set & 468 & 29791 & 1166 \\
      Test Set & 470 & 27927 & 1346 \\  \hline
    \end{tabular}
    \caption{Statistics of Named Entity Datasets}
    \label{tab:dataset} 
  \end{center}
\end{table}

\subsection*{External Retrieval}
We adopt Bing search API as an off-the-shelf search engine to retrieve external related texts. TextRank from HarvestText\footnote{\url{https://github.com/blmoistawinde/HarvestText}} and BM25 from rank\_bm25\footnote{\url{https://github.com/dorianbrown/rank_bm25}} are utilized for query generation and text similarity re-ranking.

\subsection*{Model and Training Configurations}
We use pretrained Chinese BERT provided by Huggingface\footnote{\url{https://huggingface.co/bert-base-chinese}} as the backbone structure. We compare our approach with the following two approaches,
\begin{itemize}
    \item \textbf{BERT-CRF} refers to finetuning the pretrained BERT with CRF decoder, which serves as our baseline model.
    \item \textbf{FLAT+BERT} refers to Flat Lattice Transformer \cite{li-etal-2020-flat} using BERT embedding
\end{itemize}

During training, the main hyper-parameters are shown in Table \ref{tab:hyper} 
\begin{table}[!ht]
\centering
\begin{tabular}{cc}
\hline
Item                       & Range           \\ \hline
\multicolumn{2}{c}{BERT-CRF and Ours}        \\ \hline
epochs                     & 5-10            \\
batch size                 & 12              \\
optimizer                  & AdamW           \\
bert lr                    & 3e-5            \\
fusion lr                  & 5e-3            \\
CRF lr                     & 1e-3            \\
max text length            & 128             \\
warmup                     & 0.1             \\ \hline
\multicolumn{2}{c}{Flat Lattice Transformer} \\ \hline
epochs                     & 60              \\
batch size                 & 10              \\
max text length            & 200             \\
early stop                 & 25              \\
optimizer                  & Adam            \\
lr                         & 6e-4            \\
bert lr                    & 0.05            \\
warmup                     & 0.1             \\ \hline
\end{tabular}
\caption{Hyper-parameters. We train our NER model in generic domain for 5 epochs and in social media domain for 10 epochs.}
\label{tab:hyper}
\end{table}

\subsection*{Evaluation Metrics}
We use "exact-match evaluation" to compute metrics, i.e., a named entity is considered correct only if both the span and entity type are correct. Precision, Recall and F1-score are computed on the number of true positives(TP), false positives(FP) and false negatives(TN). 
\begin{equation}
    Precision=\frac{TP}{TP+FP}
\end{equation}
\begin{equation}
    Recall=\frac{TP}{TP+FN}
\end{equation}
\begin{equation}
    F1-score = 2\times \frac{Precision \times Recall}{Precision+Recall}
\end{equation}

\subsection*{A Comparison of Fusion Strategies}
\begin{figure*}[!ht]
\centering
\subfigure[Linear Fusion Layer]{
\begin{minipage}[t]{0.5\textwidth}
\centering
\includegraphics[scale=0.25]{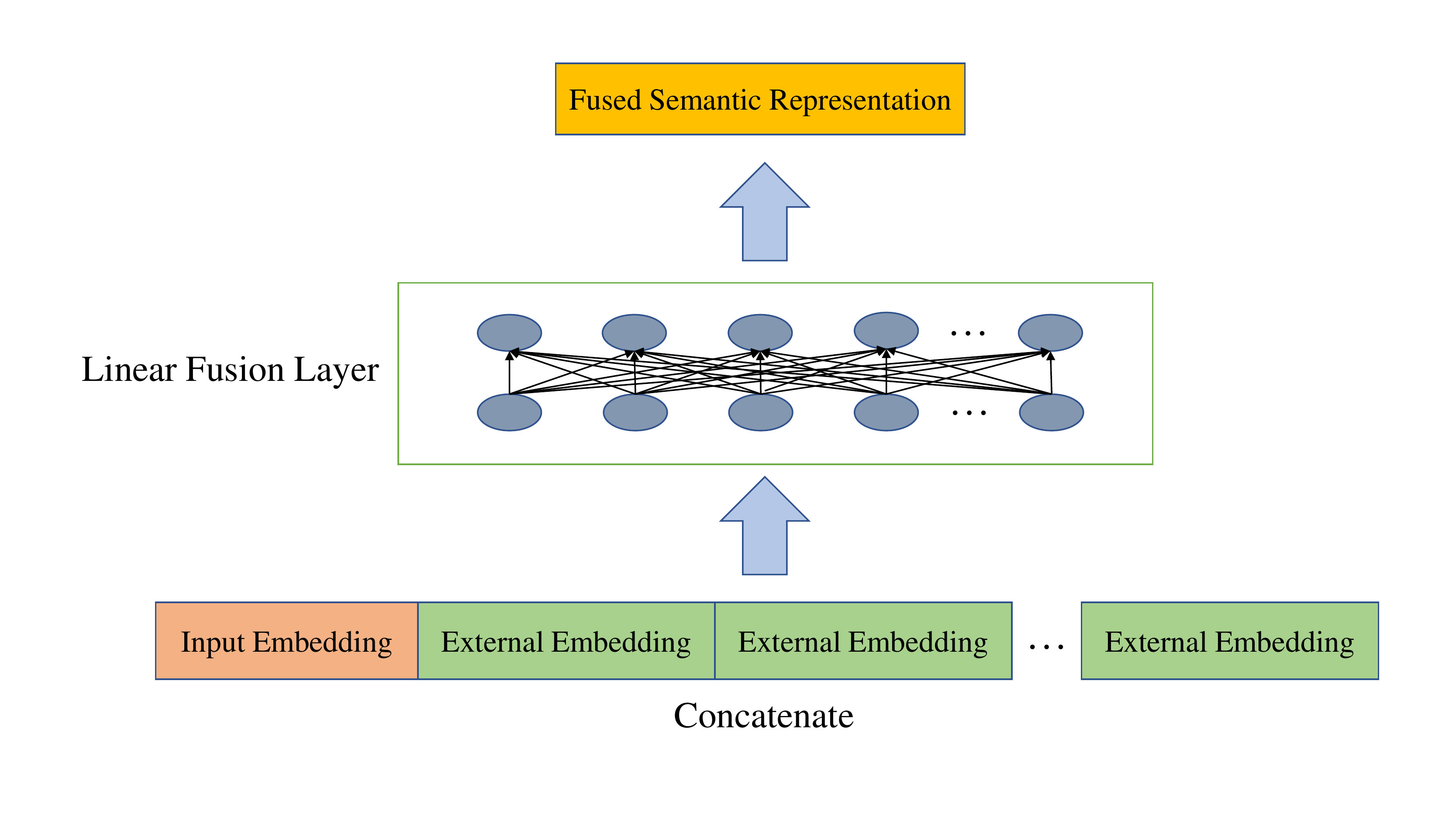}
\end{minipage}%
}%
\subfigure[Attention Fusion Layer]{
\begin{minipage}[t]{0.5\textwidth}
\centering
\includegraphics[scale=0.25]{EMNLP 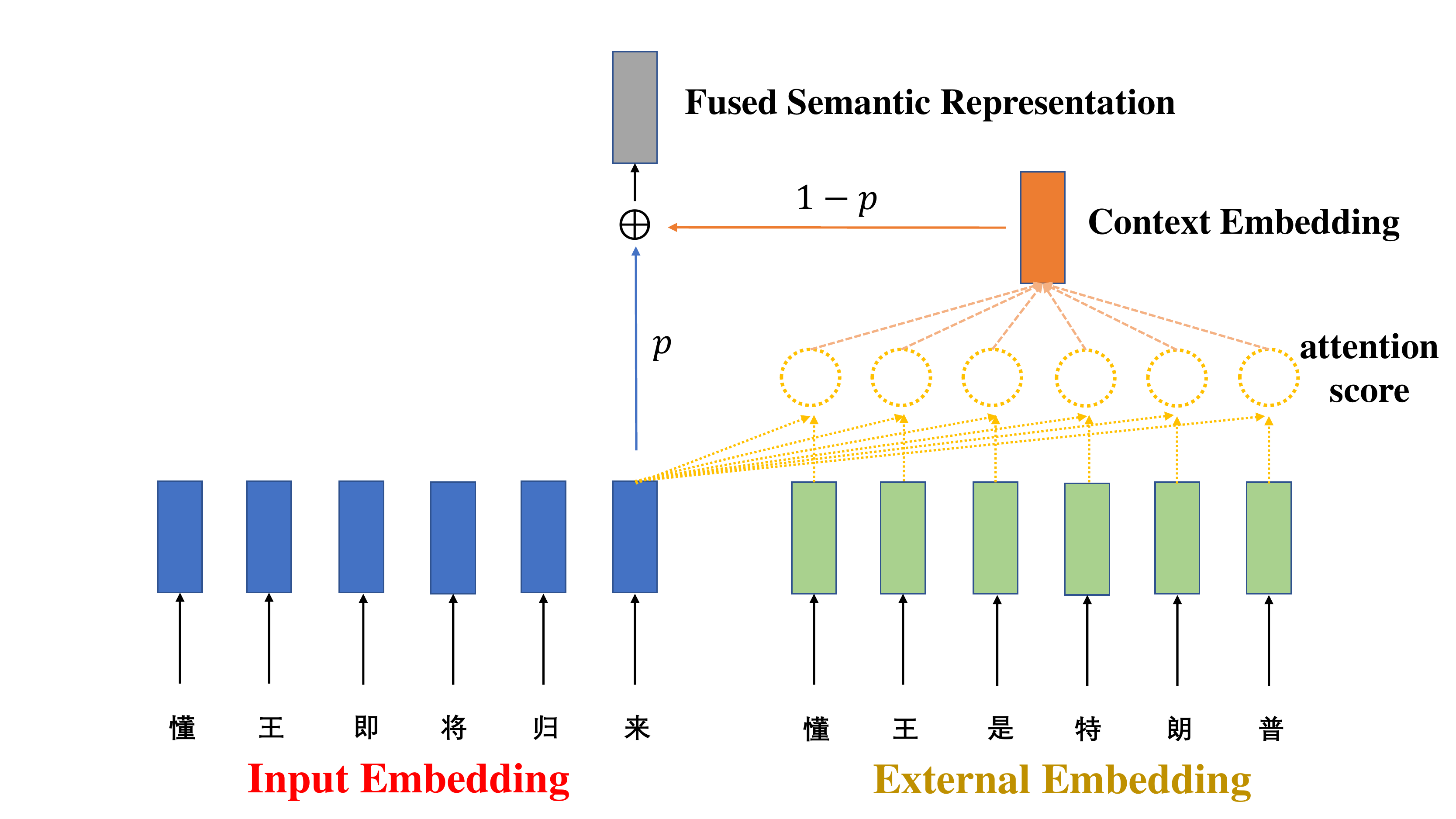}
\end{minipage}%
}%
\centering
\caption{An Illustration of Two Fusion Strategies}
\label{fig:layer}
\end{figure*}

Figure \ref{fig:layer} is an illustration of two semantic fusion strategies. A simple linear transformation is performed in the linear fusion layer, in which the input is a concatenation of input embedding and external embedding $H \in \mathbf{R}^{B \times ((K+1) \times L) \times hidden\_dim}$ and the output is $H_{fusion} \in \mathbf{R}^{B \times L \times hidden\_dim}$, where $K$ is the number of external related texts and $L$ is the max length of each text. In the attention fusion layer, we compute the attention between input embedding and external embedding and calculate the weighted sum, as described in \ref{sec:fusion}.

\begin{table*}[!ht]
    \centering
    \begin{tabular}{ccccc}
    \hline
        K & \textbf{Chinese Resume} & \textbf{People's Daily} & \textbf{Weibo NER} & \textbf{Unconventional}\\ \hline
        0 & 0.9606 & 0.9163 & 0.6905 & 0.8160 \\ \hline
        1 & 0.9571 & 0.9167 & 0.6945 & 0.8331 \\ \hline 
        2 & 0.9578 & 0.9158 & 0.6825 & 0.8318 \\ \hline
        3 & 0.9633 & 0.9184 & 0.7085 & 0.8412 \\ \hline
    \end{tabular}
    \caption{The effect of K on model performance across 4 datasets}
    \label{tab:kvalue}
\end{table*}

\subsection*{Supplementary Experiments}
\subsubsection*{Effect of External Contexts' number}
K refers to the number of external contexts which are fed into the model with original input, which represents how much external knowledge is leveraged to conduct semantic augmentation for Chinese NER. We tried different values of K ranging from 0 to 3, the result is shown in Table \ref{tab:kvalue}. Compared with F1 score for K=0, F1 score for K=3 improves by 0.27\% on Chinese Resume, 0.21\% on People’s Daily, 1.70\% on Weibo NER and 2.52\% on Unconventional. \par 
The improvement shows the effectiveness of search engine augmentation, especially for NER in social media domain. However, we find F1 scores do not always increase when K increases, a slight decrease occurs when K=1 or 2. Due to framework limitation, we do not test more values of K for now, a larger-scale experiment will be presented later.

\subsubsection*{Effect of External Knowledge's Proportion in Semantic Fusion}
\begin{figure}[ht]
\centering
\includegraphics[scale=0.45]{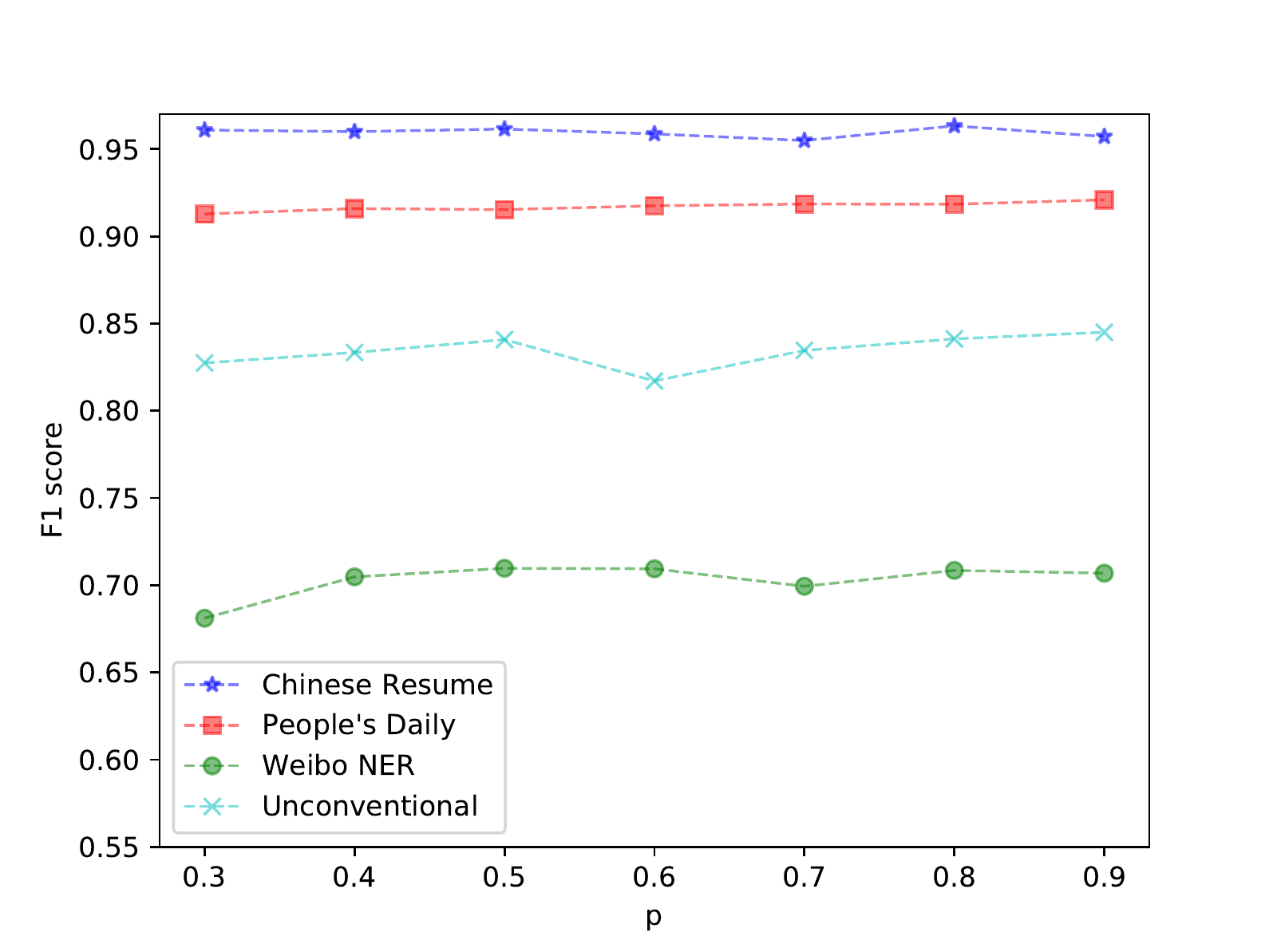}
\caption{The F1 score over different values of $p$, which represents external knowledge's proportion in semantic fusion.}
\label{fig:p_f1}
\end{figure}

Equation \ref{con:fusion_factor} shows the proportion of external knowledge in semantic fusion, in which $p$ is an adjustable parameter. We tried different values of $p$ ranging from 0.3 to 0.9 at an interval of 0.1.The result is shown in Figure \ref{fig:p_f1}. Note that F1 scores on two generic NER dataset, including Chinese Resume and People's Daily, remain stable over different $p$ values. By contrast, a flutuation of F1 scores exist on two social media dataset, including Weibo NER and Unconventional NER.\par
Figure \ref{fig:p_f1} illustrates that F1 score on Chinese Resume and People's Daily keep almost unchanged over different values of $p$, which indicates that the proportion of external knowledge has little effect on the generic NER performance. However, F1 score of social media NER does apparently depend on the proportion of external knowledge. A relationship exists between F1 score and $p$ value and it's not a simple linear relationship. Further experiments are needed to clarify this relationship.

\end{document}